\documentclass[sigconf]{acmart}
\AtBeginDocument{%
  }

\copyrightyear{2025}
\acmYear{2025}
\setcopyright{acmlicensed}
\acmConference[KDD '25] {Proceedings of the 31st ACM SIGKDD Conference on Knowledge Discovery and Data Mining V.2}{August 3--7, 2025}{Toronto, ON, Canada.}
\acmBooktitle{Proceedings of the 31st ACM SIGKDD Conference on Knowledge Discovery and Data Mining V.2 (KDD '25), August 3--7, 2025, Toronto, ON, Canada}
\acmISBN{979-8-4007-1454-2/25/08}
\acmDOI{10.1145/3711896.3737208}
\usepackage{caption}

\usepackage{listings}
\usepackage{subcaption}

\newtheorem{myDef}{Definition} 
 
\usepackage[ruled,linesnumbered]{algorithm2e}
\usepackage{enumitem}
\usepackage{multirow}
\usepackage{hyperref}
\usepackage{graphicx}
\makeatletter

\newcommand{\Rmnum}[1]{\expandafter\@slowromancap\romannumeral #1@}

\newcommand{\eg}{\emph{e.g.},\xspace}
\newcommand{\ie}{\emph{i.e.},\xspace}

\DeclareMathAlphabet\mathbfcal{OMS}{cmsy}{b}{n}

\newtheorem{pro}{Problem}

\newcommand{\eat}[1]{}
\ifodd 1

\newcommand{\TODO}[1]{{\color{red}TODO:{#1}}}

\else

\newcommand{\TODO}[1]{}

\fi

\begin{document}

\title{DiMA: An LLM-Powered Ride-Hailing Assistant at DiDi}

\author{Yansong Ning\textsuperscript{\dag}}
\affiliation{%
\institution{The Hong Kong University of Science and Technology (Guangzhou)}
\city{}
\country{}
}
\email{yning092@connect.hkust-gz.edu.cn}

\author{Shuowei Cai\textsuperscript{\dag}}
\affiliation{%
\institution{The Hong Kong University of Science and Technology (Guangzhou)}
\city{}
\country{}
}
\email{scaiak@connect.hkust-gz.edu.cn}

\author{Wei Li}
\affiliation{%
\institution{Didichuxing Co. Ltd}
\country{}
\country{}
}
\email{peterliwei@didiglobal.com}

\author{Jun Fang}
\affiliation{%
\institution{Didichuxing Co. Ltd}
\country{}
\country{}
}
\email{fangjun@didiglobal.com}

\author{Naiqiang Tan}
\affiliation{
\institution{Didichuxing Co. Ltd}
\country{}
\country{}
}
\email{tannaiqiang@didiglobal.com}

\author{Hua Chai}
\affiliation{
\institution{Didichuxing Co. Ltd}
\country{}
\country{}
}
\email{chaihua@didiglobal.com}

\author{Hao Liu\textsuperscript{\ddag}}
\affiliation{
\institution{The Hong Kong University of Science and Technology (Guangzhou)}
\city{}
\country{}
}
\email{liuh@ust.hk}
\thanks{\textsuperscript{\dag}Equal contribution. Work done during internship at Didichuxing Co. Ltd.}
\thanks{\textsuperscript{\ddag}Corresponding author.}

\renewcommand{\shortauthors}{Yansong Ning et al.}

\begin{abstract}
On-demand ride-hailing services like DiDi, Uber, and Lyft have transformed urban transportation, offering unmatched convenience and flexibility.
In this paper, we introduce \textbf{DiMA}, an LLM-powered ride-hailing assistant deployed in DiDi Chuxing. 
Its goal is to provide seamless ride-hailing services and beyond through a natural and efficient conversational interface under dynamic and complex spatiotemporal urban contexts.
To achieve this, we propose a spatiotemporal-aware order planning module that leverages external tools for precise spatiotemporal reasoning and progressive order planning. Additionally, we develop a cost-effective dialogue system that integrates multi-type dialog repliers with cost-aware LLM configurations to handle diverse conversation goals and trade-off response quality and latency. Furthermore, we introduce a continual fine-tuning scheme that utilizes real-world interactions and simulated dialogues to align the assistant's behavior with human prefered decision-making processes.
Since its deployment in the DiDi application, DiMA has demonstrated exceptional performance, achieving $93\%$ accuracy in order planning and $92\%$ in response generation during real-world interactions. Offline experiments further validate DiMA’s capabilities, showing improvements of up to $70.23\%$ in order planning and $321.27\%$ in response generation compared to three state-of-the-art agent frameworks, while reducing latency by $0.72\times$ to $5.47\times$. 
These results establish DiMA as an effective, efficient, and intelligent mobile assistant for ride-hailing services.
Our project is released at \href{https://github.com/usail-hkust/DiMA}{https://github.com/usail-hkust/DiMA} and we also release the MCP service (\href{https://mcp.didichuxing.com/api}{https://mcp.didichuxing.com/api}) to foster the ride-hailing research community. 
\end{abstract}




\maketitle

\section{Introduction}
On-demand ride-hailing services like DiDi, Uber, and Lyft have significantly transformed urban transportation systems and human mobility patterns \cite{xu2018large, lyu2024autostf, han2024bigst}. 
These platforms have introduced a paradigm shift in how people access transportation, offering unparalleled convenience, efficiency, and flexibility \cite{feng2021we, liu2022machine}. 
As one of the largest ride-hailing platforms, the DiDi mobile application processes over ten billion user requests daily, encompassing tasks such as destination selection, travel time estimation, ride-hailing, re-routing, and feedback. 
These interactions span all stages of the ride-hailing process including pre-ride, in-ride, and post-ride \cite{s2020capturing}. Despite their transformative impact, existing mobile applications still rely heavily on manual inputs like typing, clicking, and scrolling, making the process time-intensive and often inefficient \cite{hsu2023usability}.
\begin{figure}
    \centering
    \includegraphics[width=1 \linewidth]{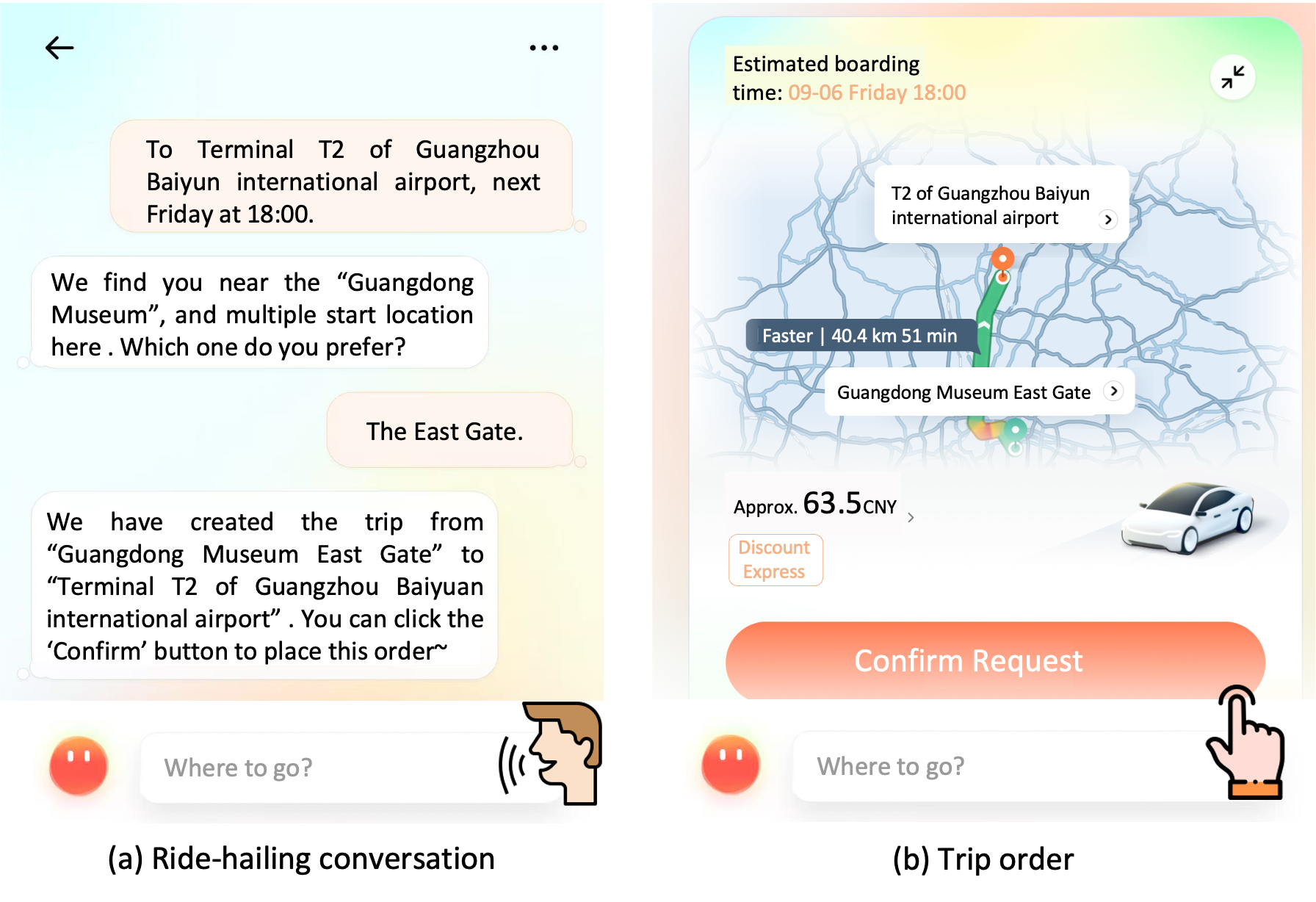}
    \caption{
    The conversational interface in DiDi Chuxing App.
    Given a user query issued on 2024-08-28, (a) DiMA proactively guides the user to complete necessary spatiotemporal trip information, and 
    (b) automatically schedules a future ride-hailing order from Guangdong Museum East Gate to Terminal T2 of Guangzhou Baiyun International Airport.
    }
    \label{fig:product}
    \vspace{-15pt}
\end{figure}

Intelligent virtual assistants such as Siri, Alexa, and Google assistant have emerged as tools to streamline user interaction by providing conversational interfaces \cite{apple_siri_guidelines}.
These intelligent assistants offer voice-activated functionality for setting reminders, searching the web, or making a phone call. 
However, their capabilities fall short in managing the unique demands of ride-hailing services, from handling open-world queries to following multi-step complex instructions that interact with the real-world urban environment.
For instance, the assistant is required to understand real-time traffic and weather conditions to accurately dispatch the order and dynamically adjust it~(\eg changing destination) during the ride. 
Recent advances in Large Language Models (LLMs), such as ChatGPT \cite{gpt35} and Qwen \cite{bai2023qwen}, have introduced new possibilities for creating domain-specific intelligent assistants. 
LLMs demonstrate exceptional capabilities in understanding and processing complex natural language queries and have been adopted to construct autonomous web operation agents. 
For example, Mind2Web \cite{deng2024mind2web} and AutoWebGLM \cite{lai2024autowebglm} serve as human-like web navigation assistants, while LLMPA \cite{guan2023intelligent} automates multi-step travel planning tasks in Alipay. Inspired by these attempts, we propose constructing an LLM-powered assistant to enable natural, conversational ride-hailing for the DiDi application.

\begin{figure}
 
    \centering
    \begin{subfigure}[b]{\linewidth}
       
        \centering
        \includegraphics[width=\linewidth]{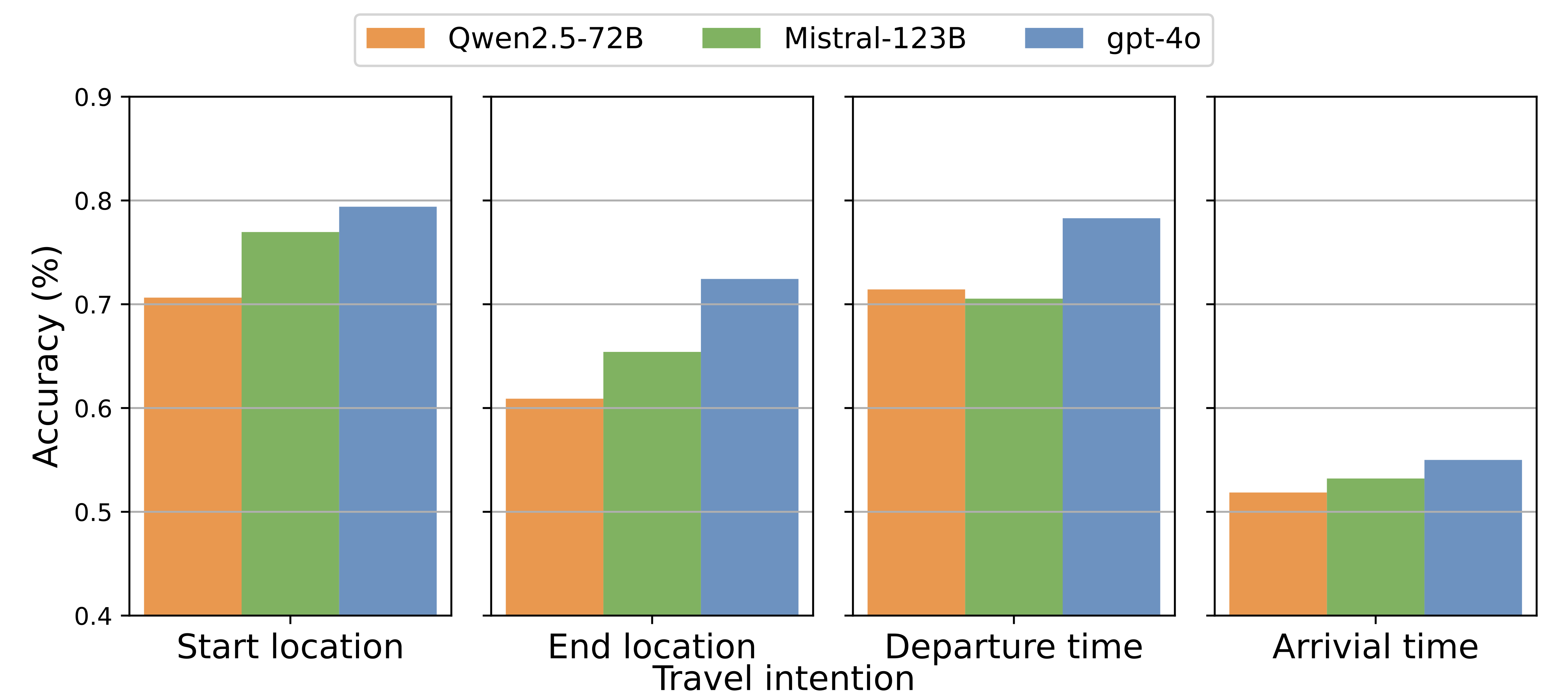}
        \caption{Existing LLMs fail to meet the production accuracy requirements on travel intention understanding.}
        \label{fig:intro_a}
    \end{subfigure}

    \begin{subfigure}[b]{\linewidth}
       
        \centering
        \includegraphics[width=\linewidth]{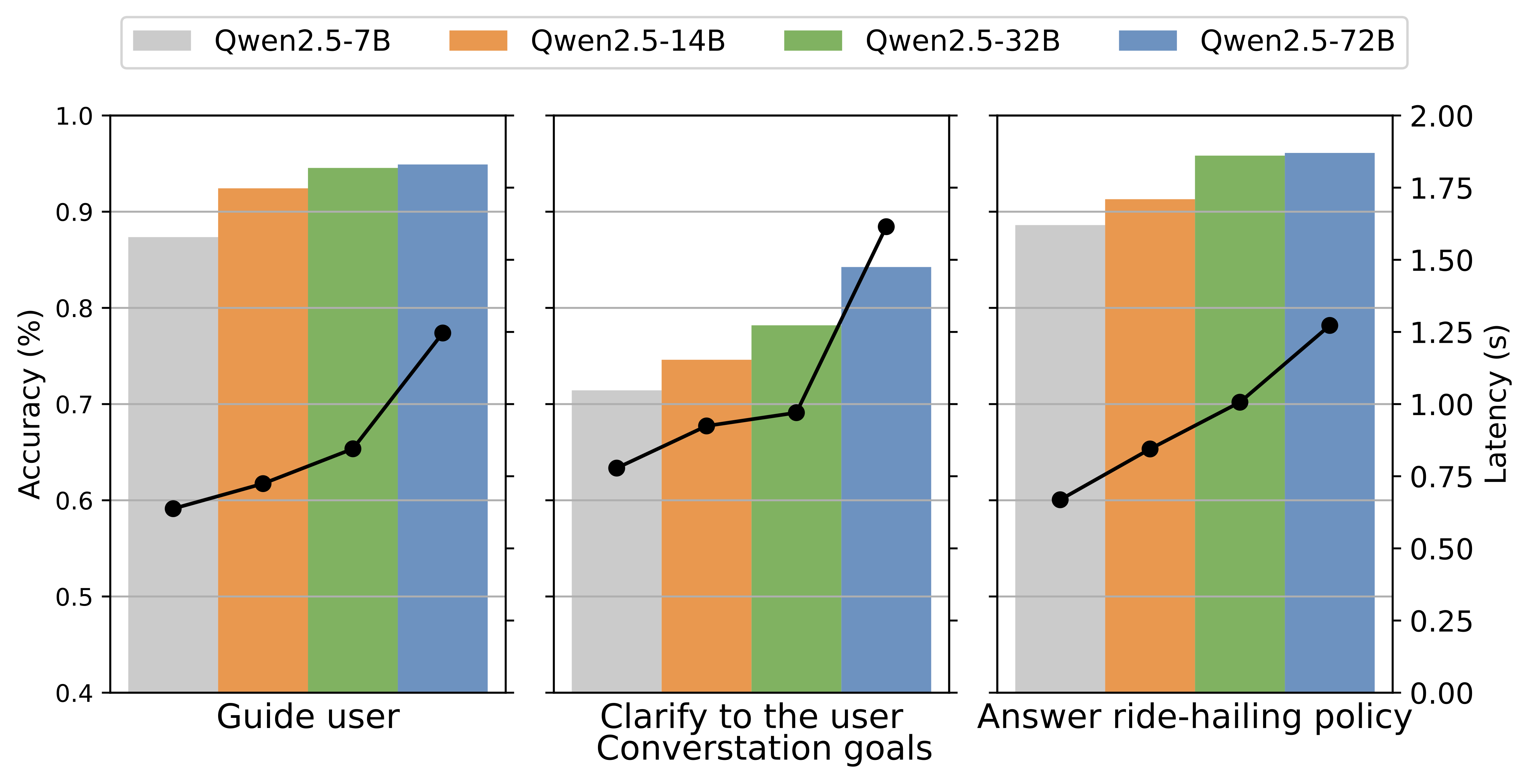}
        \caption{Quality and latency trade-off of size-varying LLMs for response generation. Under three conversation goals, larger LLMs tend to generate better responses over a longer inference time.}
        \label{fig:intro_b}
    \end{subfigure}
    \caption{Performance of existing LLMs as a ride-hailing assistant. The results are obtained based on 500 human annotated real-world ride-hailing requests.}
    \label{fig:intro}
\vspace{-15pt}
\end{figure}

However, building an LLM-based ride-hailing assistant is a non-trivial task due to the following three challenges.
\emph{(1) Spatiotemporal Travel Intention Understanding.} The ride-hailing process relies heavily on understanding the spatiotemporal trip information \cite{wang2024language} such as destination and departure time. 
LLMs often struggle with spatial reasoning \cite{xie2024travelplanner} and temporal calculations \cite{su2024timo}. 
For instance, when issuing a query shown in Figure \ref{fig:product}, general LLMs may fail to derive the exact date of next Friday and neglect the user's current location as the start location.
As reported in Figure \ref{fig:intro}(a), our quantitative studies validate that even state-of-the-art LLMs like GPT-4o and Qwen2.5-72B achieve less than 80\% accuracy in spatiotemporal travel intention understanding, therefore limiting their applicability for handling real-world ride-hailing requests. 
\emph{(2) Proactive Order Planning.} Ride-hailing involves multiple factors (\eg waiting time, willing car type, destination) for matching passengers and drivers, which may require multi-turn conversations. 
Moreover, during a conversation, users may cancel confirmed orders, request orders with infeasible conditions (\eg unrealistically low costs), or engage in ride-hailing policy inquiry. 
How to proactively and naturally guide users through the order planning process also poses a significant challenge.
\emph{(3) Cost-Effective Response Generation.} The assistant should generate responses adapted to varying conversation goals and order statuses. 
For instance, guiding users to place an order often requires formulaic language responses, handling infeasible request needs clarification, whereas answering ride-hailing policy queries may involve retrieving DiDi’s knowledge base.
As shown in Figure \ref{fig:intro}(b), in general, larger LLMs tend to be more versatile to handle complex scenarios but result in higher response latency. Generating accurate responses within strict online service latency is another critical challenge.

To address the aforementioned challenges, we propose \textbf{DiMA}, an LLM-based intelligent assistant for ride-hailing at DiDi.
As its core, we first propose a \emph{Spatiotemporal-aware Tool-augmented Order Planning} module to enhance LLMs’ ability to interpret spatiotemporal queries by leveraging external tools for precise location mapping and temporal reasoning. By injecting domain knowledge and utilizing tools like time calculators and geographic databases, DiMA accurately extracts travel intentions (\eg start and end locations and preferred departure times), facilitating progressive order planning.
Moreover, we construct a \emph{Cost-effective Dialog System} to handle diverse conversation goals in a cost-effective way. By proactively guiding users through multi-type dialog repliers, the dialog system ensures seamless order planning and addresses diverse scenarios, from complex order planning to ride-hailing policy answering. The system further adopts a cost-aware LLM configuration, which adaptively allocates smaller models for simpler queries and larger models for complex scenarios, ensuring high-quality responses while meeting strict production latency requirements.
Finally, we propose a \emph{Continual Fine-tuning} scheme to align the assistant with human decision processes. By continuously collecting real-world user interactions and augmenting user decision trajectories from a constructed role-playing simulation environment, we periodically fine-tune backbone LLMs to overcome cold start problem and align the assistant behaviors with shifting user preferences, thereby enabling proactive and natural order planning guidance.

Since May 2024, DiMA has been deployed in the DiDi application, as depicted in Figure \ref{fig:product}. 
Two months’ online experiments in production demonstrate that DiMA achieves 93\% accuracy in order planning and 92\% in response generation. 
We further conduct offline experiments on synthetic datasets in Beijing and Shanghai. The results validate that DiMA significantly outperforms three state-of-the-art agent frameworks on order planning and response generation tasks, with improvements up to 70.23\% and 321.27\%, respectively.
Moreover, the results show that DiMA maintains comparable performance to models like Qwen2.5-72B, Mistral-123B, and Llama3.1-405B, while reducing latency by 0.72x to 5.47x.

Our contributions are summarized as follows: (1) We propose the first LLM-based assistant deployed in DiDi mobile application, offering a new interaction alternative for ride-hailing; (2) We develop a spatiotemporal-aware tool-augmented order planning module and a cost-effective dialogue system to enhance accuracy and efficiency; (3) We introduce a continual fine-tuning scheme to ensure sustained proactive order planning guidance aligned with human decision processes; (4) Extensive experiments validate DiMA’s effectiveness and demonstrate its advantages against state-of-the-art LLM agent frameworks in both online and offline settings.

\section{Preliminaries}

\subsection{Definition and Problem Statement}
We begin with the definition of trip order and ride-hailing conversation, then define the problem we aim to address.

\begin{myDef}
\textbf{Trip order.}
A trip order is defined as a 9-tuple $o = \left (l_{s}, l_{v}, l_{e}, t_{d}, t_{a}, dist, dur, c, p\right )$, where $l_{s}$, $l_{v}$ and $l_{e}$ denote the start location, via location and end location of the trip, $t_{d}$ and $t_{a}$ is the trip departure time and estimated arrival time. In addition, $dist$ and $dur$ denote the distance and estimated duration of the trip, $c$ is the car type, and $p$ is the expected price of this trip.
\end{myDef}

Note that the start location $l_{s}$, via location $l_{v}$, and end location $l_{e}$ usually corresponds to Point-of-Interests (POIs) associated with geographical coordinates. 
Figure \ref{fig:product}(b) illustrates a trip order example.

\begin{myDef}
\textbf{Ride-hailing conversation.}
Let $q_i$ and $r_i$ denote the user query and assistant response at round $i$,
a conversation $d$ is defined as a sequence of queries and responses $\{q_{1}, r_{1}, q_{2}, r_{2}, ...\}$.
In the context of ride-hailing, a query $q_i$ can be a trip order request, a policy inquiry, or even a chit-chat. 
$r_i$ can be responses proactively guiding the user to provide specific instructions, \eg providing preferences, confirming the order, and re-routing, or also be responses answering questions, providing travel suggestions, etc.

\end{myDef}

Note in a conversation, a trip order $o$ can be completed by multi-turn user queries. 
Figure \ref{fig:product}(a) show a conversation example.


\eat{
\begin{pro}
Given a user query $q_{i}$ at round $i$, we aim to create the trip order $o_{i}$ and provide corresponding response $r_{i}$ based on history dialog $\mathcal{H} = 
\left \{  q_{1}, r_{1}, q_{2}, r_{2}, ..., q_{i-1}, r_{i-1} \right \}$, which is a set of historical user queries and assistant responses from round $1$ to round $i-1$.
\end{pro}
}

\begin{pro}
We aim to construct an end-to-end mobile assistant for seamless ride-hailing services through a conversational interface, i.e. ride-hailing conversation.
As its core, the assistant automatically guides the users to provide the required spatiotemporal trip information, reasoning about user travel intentions under real-world urban contexts, and creating the trip order $o$ to satisfy the trip requirements.
 
\end{pro}

\subsection{The Ride-Hailing Service at DiDi}
The ride-hailing service at DiDi continuously matches passenger requests with available drivers \cite{xu2018large}. The service can be structured into three key stages: pre-ride, in-ride, and post-ride \cite{s2020capturing}.

In the pre-ride stage, users interact with the application by setting their origin and destination, scheduling the departure time, and selecting their preferred car type. The app then provides essential details, such as estimated arrival times \cite{liu2022machine, liu2023uncertainty} and expected prices. Once the user confirms the order, the process transitions into the in-ride stage, where users can make route changes, update prices, and access additional ride details. Finally, in the post-ride stage, users complete payment and are encouraged to provide feedback, which helps improve service quality.

Despite the diverse goals of the ride-hailing service, interactions primarily rely on manual inputs such as typing, clicking, and scrolling, which are time-consuming and inefficient. To enhance the user experience, DiDi has tried integrating Task-oriented Dialogue (ToD) systems \cite{qin2023end}, which simplify the ride-hailing process through natural language interaction. However, traditional ToD methods \cite{hosseini2020simple, he2022galaxy} are typically designed for fixed tasks, such as reservations or question answering, which often fall short of understanding the spatiotemporal travel context and conduct open-world reasoning. These limitations make ToD an impractical solution for real-world ride-hailing scenarios where user intents are often dynamic. For instance, users may wish to change destinations in-ride, inquire about estimated arrival times, or engage in casual chit-chat.

Based on the above observations and attempts, next we introduce our proposed LLM-based ride-hailing assistant framework, designed to handle dynamic user interactions and adapt to the complexities of real-world ride-hailing scenarios.
\begin{figure*}   
    \centering
    \includegraphics[width=1 \linewidth]{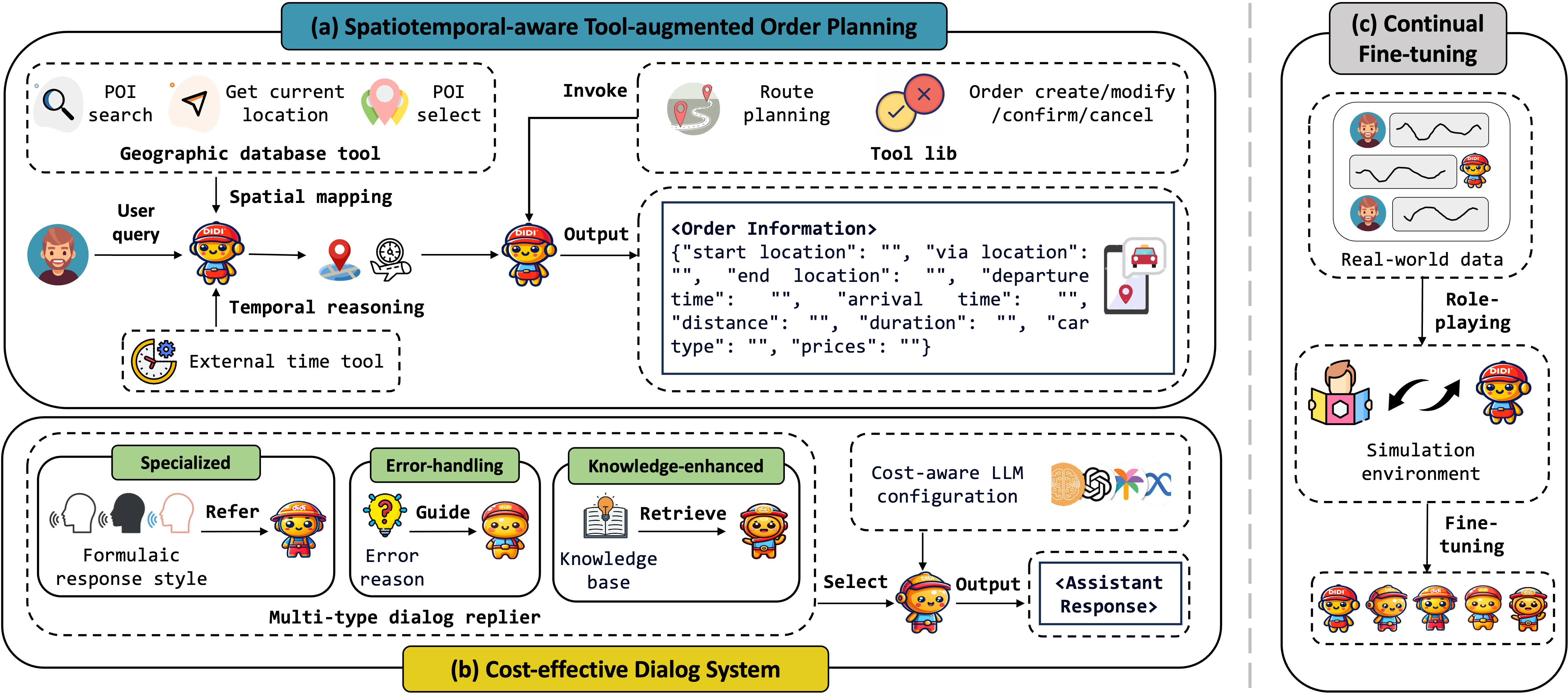}
    \caption{An overview of DiMA.
    }
    \label{fig:method}
\end{figure*}

\section{Framework Overview}
Figure \ref{fig:method} shows the framework of DiMA, which is comprised of three major components, \emph{Spatiotemporal-aware Tool-augmented Order Planning}, \emph{Cost-effective Dialog System} and \emph{Continual Fine-tuning}.
Given an online user query, \emph{Spatiotemporal-aware Tool-augmented Order Planning} first conducts spatial mapping and temporal reasoning to extract travel intention (\ie start, via and end location, departure and arrival time), and then executes progressive order planning to obtain order information.
Moreover, the \emph{Cost-effective Dialog System} constructs multi-type dialog replier to handle diverse conversation goals.
Based a cost-aware LLM configuration, the system adaptively allocate smaller models for simpler queries and larger models for complex queries, improving the response quality and reduces the latency.
Besides, the \emph{Continual Fine-tuning} schema periodically fine-tunes models using real-world ride-hailing data from the mobile application together with instructions acquired from the constructed role-playing simulator to align the assistant behavior with the user decision-making process.

\section{Spatiotemporal-Aware Tool-Augmented Order Planning}
One key objective of DiMA is to achieve accurate order planning.
Over time, DiDi engineers have developed a comprehensive suite of spatial and temporal tools that encapsulate extensive knowledge and operational insights, which can be leveraged to enhance the assistant’s spatiotemporal reasoning and multi-step planning capabilities. 
In this section, we introduce the \emph{Spatiotemporal-aware Tool-augmented Order Planning} module, including spatial mapping, temporal reasoning, and progressive order planning.

\subsection{Spatial Mapping}
In ride-hailing scenario, user queries usually contain crucial spatial travel information, which is essential for accurate order planning. As reported in Figure \ref{fig:intro}(a), general LLMs lack specialized capabilities to effectively extract and interpret spatial travel intention. To address this limitation, we enable DiMA to learn from and utilize spatial tools through structured function calls~\cite{li2024large}.

Specifically, we establish a structured pipeline encompassing four key stages: 
(1) Spatial tool collection and annotation. 
We first collect tool APIs from DiDi’s code base that related to spatial information processing and analysis, such as \emph{POI\_search()} and \emph{Get\_current\_location()}. 
These APIs are carefully annotated by engineers, including metadata such as name, description, required parameters, optional parameters, and expected outputs. 
A pool of geographic database tool is maintained for spatial mapping.
(2) Spatial decomposition. 
Based on historical ride-hailing interactions and operational needs, we define a set of structured spatial tasks, such as extracting start and end location from the user query, completing missing locations. 
The entire process is structured as a sequence of tool interactions.
(3) Tool selection. Given a user query, DiMA first classifies the required process and dynamically selects required tool from the pool based on spatial annotations.
(4) Tool calling. Once the appropriate tools are selected, DiMA generates structured function calls \cite{li2024large} by extracting and completing key parameters.

For example, given the user query shown in Figure \ref{fig:product}, where neither the start nor end location is explicitly specified.
DiMA defaults to treating the mentioned location as the destination, and automatically generate structured function call \emph{POI\_search("Terminal T2 of Guangzhou Baiyuan international airport")} to determine the destination and using \emph{Get\_current\_location()} to infer the starting location. 
After calling corresponding geographic database tools, the extracted spatial intention is mapped to a real-world POI, \eg [\emph{"Terminal T2 of Guangzhou Baiyuan international airport", 23.3672, 113.2998, 2506217808}].
Finally, we derive the mapped start/via/end locations, which will be passed for order planning.

\subsection{Temporal Reasoning}
Besides spatial mapping, ride-hailing interactions also require accurate temporal reasoning, \eg understanding of departure and arrival times. As illustrated in Figure \ref{fig:intro}(a), LLMs struggle with accurate date and time interpretation, achieving less than 80\% and 60\% accuracy on departure and arrival time understanding, respectively.

Similar to spatial mapping, we construct a four stage pipeline: (1) Temporal tool collection and annotation, (2) Temporal decomposition. (3) Tool selection, and (4) Tool calling.
The whole process is similar with spatial mapping.
Take the user query in Figure \ref{fig:product} again for example.
Since the user's departure time may not be explicitly specified, we follow product design to threat mentioned time as departure time.
Therefore, DiMA will automatically generate structured function call \emph{Get\_departure\_time()} with key parameters \{\emph{"date": "1 week later, Fri", "time": "18:00:00"}\} to obtain user's departure time.
Then, based on the issued date \emph{"2024-08-28"}, DiMA calls time tool to transform it into the exact datetime [\emph{"2024-09-06 18:00:00"}].

\subsection{Progressive Order Planning}
\label{Progressive Order Planning}
With the extracted spatiotemporal intentions, DiMA executes progressive order planning through tool-assisted decision-making.
However, ride-hailing interactions often require multi-turn conversations, where each round corresponds to refining or supplementing spatiotemporal intentions. 
Additionally, ride-hailing requests encompass diverse conversation goals, including order creation, modification, confirmation, cancellation, ride-hailing policy inquiries, or casual chit-chat. 
Recognizing and responding dynamically to multi-turn user interactions is critical for ensuring a seamless and responsive ride-hailing experience.

\begin{figure}
\centering
\begin{lstlisting}
def Get_current_location(session_id):
    POI_list = requests.post(session_id)
    # [{displayname, lat, lng, id}, ...]
    return POI_list

def POI_search(session_id, POI_name):
    POI_list = requests.post(session_id, POI_name)
    return POI_list

def POI_select(POI_list, Selected_POI_name):
    return Selected_POI
    
def Route_planning_API(session_id, SrcLat, SrcLng,
ViaLat, ViaLng, DstLat, DstLng, SrcId, ViaId, DstId):
    order_list = requests.post(session_id, ...)
    return Order_list

def Time_tool(date, time, current_time):
    """
    date: "%Y-%m-%d" or "N week later, Mon" or "absent"
    time: "%H:%M:%S" or "absent"
    current_time: "%Y-%m-%d %H:%M:%S"
    """
    return date_time
\end{lstlisting}
\caption{API and tool lib.}
\label{apis and tools}
\vspace{-15pt}
\end{figure}
To achieve this, we formulate the order planning process as structured functions (\eg \emph{Order\_create()} and \emph{Order\_cancel()}), which takes multiple spatial and temporal input parameters. 
The assistant learns to invoke these functions by proactively seeking missing input values—such as start/via/end locations and departure/arrival times—through multi-turn dialog. 
For instance, in the illustrative example, executing tools like \emph{POI\_search()} results in structured travel intentions, including extracted start and end locations. 
This information is subsequently processed via \emph{Route\_planning()} to retrieve estimated trip distance, duration, vehicle type options, and pricing details. 
Once all parameters are determined, the assistant invokes \emph{Order\_create()} to obtain the order information.

Furthermore, we construct illustrative spatiotemporal descriptions for various conversation goals and integrate them into the LLM prompt. 
For example, during policy inquiries, users typically do not specify structured spatial or temporal information, distinguishing these interactions from explicit ride-hailing requests. 
By leveraging illustrative description guidance, the assistant enhances its instruction-following ability to execute complex order planning, such as confirming, modification, or canceling an order.

In this way, the assistant ensures reliable spatiotemporal interpretation and order planning, making it well-suited for real-world ride-hailing scenarios. Figure \ref{apis and tools} shows tool information and please refer to Appendix \ref{prompt_template} for more detailed prompt template.

\section{Cost-Effective Dialog System}
Then, we present the \emph{Cost-effective Dialog System}, including multi-type replier construction and cost-aware LLM configuration, designed to optimize assistant responses to fulfill the production requirements. 
These components enable the dialog system to generate high-quality responses with reduced latency while adapting to various conversation goals, such as order creation, modification, policy inquiries, and casual conversations.

\subsection{Multi-Type Replier Construction}
In online serving, the response generation policy plays a crucial role in determining how the assistant interacts with users~\cite{young2013pomdp}.
Following DiDi's product design, we decompose dialog policy into three distinct types and construct corresponding LLM repliers.
\textbf{(1)~Specialized replier:} This replier is designed for cases where the order can be properly created with alignment of the user’s intention. 
It follows a structured, formulaic~\cite{wray2012we} response style to proactively guide the users' action through ride-hailing, such as prompting them to confirm the created order. This replier is predominantly used for order planning tasks.
\textbf{(2)~Error-handling replier:} 
Not all user requests can be satisfied.
Handling infeasible or conflicted user requests and accidental system errors are essential for maintaining a smooth user experience \cite{bohus2005error}. 
This replier intervenes when an order request cannot be fulfilled or when incorrect information is detected. 
Despite rarely invoked in practice, it generates clarification messages or apologetic responses to help users refine their requests and prevent frustration.
\textbf{(3)~Knowledge-enhanced replier:} In cases where users inquire about company policies or engage in casual conversations (\eg ``How many car types can I book now?'' and ``What is the traffic restriction policy today?''), this replier retrieves relevant information from DiDi’s dynamically maintained knowledge base~\cite{gao2023retrieval}. Retrieved question-answer pairs are appended to the LLM prompt, ensuring the response is factually accurate and contextually relevant.

In practical implementation, we provide the description of each dialog policy in the prompt to guide DiMA select the most suitable replier for response generation.
To enhance response accuracy, for each dialog replier, we provide task descriptions along with demonstration examples in the prompt. 
Detailed prompt structures can be found in Appendix \ref{prompt_template}.

\subsection{Cost-Aware LLM Configuration}
Given the selected dialog replier, using an appropriate language model for response generation requires balancing computational efficiency and performance. A naive approach would always be to employ a high-capacity LLM (\eg Qwen2-72B) to ensure ride-hailing response quality. However, this approach is impractical due to system budget and latency considerations~\cite{sarikaya2016overview, padmanabha2018mitigating}. Instead, we propose a cost-aware LLM configuration strategy.

Our key insight is that for each utterance, some dialog goals~(\eg policy answering) are simple and can be handled by smaller, generalist LLMs, whereas others~(\eg error-handling) require customized reasoning capabilities of larger models. Along this line, we first construct an LLM pool containing both high-latency but strong LLMs and low-latency but weaker ones. 
Following~\cite{bansal2024smaller, chen2024self}, we carefully measure the cost-effectiveness of each language model variant for each replier and choose the best configuration (\ie, model family and size) for response generation. More implementation details will be discussed in Section~\ref{deployment}. Through these optimized replier customizations, we adaptively allocate LLM resources based on the given dialog policy. This approach ensures high system throughput while preserving response quality.

\section{Continual Fine-Tuning}
\label{Continual Fine-tuning}

This section details \emph{Continual Fine-tuning}, aiming to align assistant behaviors with user preferences, enabling proactive and natural ride-hailing guidance and beyond.
By collecting real-world user dialogs and synthesizing dialogs based on multi-turn user interaction trajectories through role-playing simulation, we continuously fine-tune the model to improve the assistant effectiveness.

\subsection{Data Collection and Augmentation}
\label{role-playing environment}
We collect real-world ride-hailing dialogs from online user requests, which are periodically stored in a MySQL database. Each record corresponds to a multi-turn conversation between a user and the ride-hailing assistant, \eg \emph{<user: Hi, book a ride to Tsinghua for me; assistant: Sure; user: ...; assistant: ...>}.

Since real-world ride-hailing interactions were initially scarce and lacked sufficient scenario coverage, we further augment the dataset using a role-playing simulator. 
Specifically, we first sample historical trip orders $\mathbf{O}=\{o_1, o_2, \dots \}$ and ride-hailing conversations $\mathbf{D}=\{d_1, d_2, \dots \}$ from DiDi's application database. 
Each record is associated with a user profile (\eg age, sex, occupation). To protect user privacy and enhance dataset diversity, user profiles are synthetically generated based on a predefined schema rather than real-world user information.
Using the above information, we employ GPT-4o to simulate both passenger and ride-hailing assistant roles, generating structured multi-turn dialogs that mimic real-world ride-hailing interactions. For example, a generated conversation might include: \emph{<user: Hi, book a ride to Tsinghua for me; assistant: Sure; user: ...; assistant: ...>}.
To ensure the quality of simulated dialogs, we employ a GPT-based evaluation mechanism \cite{zheng2023judging} to filter out unfaithful or low-quality samples before constructing the final instruction tuning dataset. 
The detailed prompt templates used for dialog role-playing simulation and instruction dataset statistics are provided in Appendix \ref{simulation}.

\subsection{Continuous Model Tuning}
Since real-world ride-hailing dialogs are continuously collected from the online application and unexpected failure cases may emerge over time, DiMA employs an automated fine-tuning pipeline to iteratively optimize the assistant. 
Let $\mathcal{D}_{1}$, $\mathcal{D}_{2}$, ..., $\mathcal{D}_{t}$ represent the mixed instruction sets constructed above, where the subscript indicates different time steps, and let $\mathcal{M}$ denote the backbone language model. The probability distribution of response $y$ given instruction $x$ is represented as $\mathcal{P}_{\mathcal{M}}\left ( y \mid x \right ) $. 
The continual tuning process is formulated as minimizing the time-evolving loss:

\begin{equation}
\label{fine-tune}
\begin{aligned}
\mathcal{L}_{t} = \mathbb{E}_{(x, y) \sim \mathcal{D}{t}}\left[\log \mathcal{P}_{\mathcal{M}}(y \mid x)\right],
\end{aligned}
\end{equation}
where $\mathcal{L}_{t}$ represents the loss function at a given time step $t$. We utilize the Low-Rank Adaptation (LoRA)~\cite{hu2021lora} to efficiently fine-tune DiMA with minimal computational overhead. This process allows DiMA to evolve dynamically, ensuring it remains aligned with the latest user interactions and ride-hailing trends.

\section{System Deployment}
\label{deployment}
Since May 2024, DiMA has launched on the DiDi mobile application to provide intelligent ride-hailing assistant service.
In this section, we present the deployment experience of DiMA, including the model selection and adaptation mechanism, asynchronous model update, and online serving strategy.

\textbf{Model selection and adaptation.}
The order planning and response generation module requires size-varying LLMs to achieve cost-effective ride-hailing service.
Specifically, we fine-tune Qwen2-72B and Qwen2-7B for spatiotemporal-aware tool-augmented order planning and multi-type dialog replier selection.
For response generation, we fine-tune Qwen1.5-32B for the specialized replier.
Besides, we observe fine-tuning leads to performance degradation for both error-handling generation and knowledge retrieval. Therefore, we choose vanilla Qwen-2-72B and Qwen-1.5-32B for error-handling and knowledge-enhanced repliers.
The above configuration guarantees the effectiveness and efficiency of DiMA cost-effectively.

\textbf{Asynchronous model update.}
As described in Section~\ref{Continual Fine-tuning}, the LLMs are continuously updated to enhance the ride-hailing assistant performance.
Benefiting from the decoupled LLM configuration for each module, we devise an asynchronous model update mechanism. 
Specifically, we employ a professional annotation team to review and label ride-hailing data collected from the online environment.
Based on the identified failure cases, we periodically refine prompts and fine-tune each model.
Once product managers validate the updated model, the latest version will be pushed into the online environment.

\textbf{Online serving.}
To optimize the online inference latency, we apply Activation-aware Weight Quantization (AWQ) techniques \cite{lin2024awq} to compress and optimize LLMs for deployment.
Additionally, we integrate the Medusa acceleration framework~\cite{cai2024medusa} to efficiently deploy LLMs of different sizes: a 72B-size model on 4 NVIDIA H800s, a 32B-size model on 2 NVIDIA H800s, and a 7B-size model on a single NVIDIA H800. This deployment strategy ensures optimal resource utilization while maintaining high responsiveness.
By incorporating these techniques, we successfully reduce DiMA’s average response latency within 4 seconds, ensuring both efficiency and scalability in real-world ride-hailing applications.

\section{Experiments}
We conduct extensive experiments to evaluate the proposed method.
We aim to answer the following research questions:
\textbf{RQ1:} How does DiMA perform in the online environment?
\textbf{RQ2:} How does DiMA perform compared with existing agent frameworks?
\textbf{RQ3:} How do different components affect the assistant's performance?
\textbf{RQ4:} How does continual fine-tuning schema enhance DiMA?
\textbf{RQ5:} How about the cost-effectiveness of DiMA?

\begin{table}[]
\caption{The statistics of online dataset, and two offline datasets in the role-playing dialog simulation environment.}
\label{dataset}
\begin{tabular}{c|ccc}
\midrule
\multirow{2}{*}{Data type}  & \multirow{2}{*}{Online} & \multicolumn{2}{c}{Offline} \\
                            &                         & Beijing       & Shanghai      \\ \midrule
Time span                   & \multicolumn{3}{c}{2024/05/06 - 2024/08/24}             \\
\# of training user queries & \multicolumn{3}{c}{2,086}                               \\
\# of testing users                 & 3,624 &    272 &  168                         \\
\# of testing user queries  & 10,153                  & 1,139         & 687           \\ 
\# of average dialog rounds  & 2.82                  & 4.19        & 4.08           \\ \midrule
\end{tabular}
\end{table}

\begin{table*}[]
\caption{Online experimental results on DiDi Chuxing ride-hailing platform.}
\label{online results}
\scalebox{0.85}{
\begin{tabular}{cc|cccc|cccc}
\midrule
\multicolumn{1}{c|}{\multirow{2}{*}{User query types}} & \multirow{2}{*}{Percentage} & \multicolumn{4}{c|}{Human Evaluation($\%$)} & \multicolumn{4}{c}{GPT Evaluation($\%$)} \\
\multicolumn{1}{c|}{}                                 &                             & ROA       & RRA       & SOA      & SRA      & ROA      & RRA      & SOA      & SRA     \\ \midrule
\multicolumn{1}{c|}{Order / modify / confirm / cancel order} & 89.26\%                     & 93.48     & 91.34     & 88.15    & 85.37    & 86.63    & 94.09    & 72.09    & 86.84   \\
\multicolumn{1}{c|}{Policy inquiries}             & 2.07\%                      & 90.02     & 97.31     & 88.89    & 94.32    & 83.76    & 98.18    & 78.36    & 95.52   \\
\multicolumn{1}{c|}{Casual chit-chat}                        & 8.67\%                      & 98.62     & 98.61     & 96.36    & 96.35    & 98.72    & 97.19    & 98.26    & 96.58   \\ \midrule
\multicolumn{2}{c|}{Overall}                                                        & 93.83     & 92.17     & 87.11    & 85.02    & 86.94    & 94.66    & 70.32    & 86.68   \\ \midrule
\end{tabular}
}
\end{table*}

\begin{table*}[]
\caption{Overall performance of order planning and response generation task on simulation environment on Beijing and Shanghai dataset. 
The best baselines performance is \underline{underlined}.}
\label{simulation results}
\scalebox{0.85}{
\begin{tabular}{cc|cccccccc|cccccccc}
\midrule
\multirow{3}{*}{Backbones} & \multirow{3}{*}{Methods} & \multicolumn{8}{c|}{Beijing}                                          & \multicolumn{8}{c}{Shanghai}                                         \\
                           &                         & \multicolumn{4}{c|}{Human Evaluation ($\%$)}                 & \multicolumn{4}{c|}{GPT Evaluation ($\%$)} & \multicolumn{4}{c|}{Human Evaluation ($\%$)}                 & \multicolumn{4}{c}{GPT Evaluation ($\%$)} \\
                           &                         & ROA & RRA & SOA & \multicolumn{1}{c|}{SRA} & ROA  & RRA & SOA  & SRA & ROA & RRA & SOA & \multicolumn{1}{c|}{SRA} & ROA  &  RRA & SOA & SRA \\ \midrule
\multirow{3}{*}{Qwen-2-72B} & CoT                    & 56.49    &76.57     &17.65     & \multicolumn{1}{c|}{45.59}    &  61.24    &  50.02    &  14.26    & 6.62   & 50.75    &  68.66   & 14.29    & \multicolumn{1}{c|}{42.86}    &  59.12     & 48.14    &  13.81   &  8.33  \\
                           & ReAct                  &  51.93   &  77.25   & 14.71    & \multicolumn{1}{c|}{45.59}    & 63.26    &  51.44  &  16.47    & 6.99   &  48.28   & 71.72    & 23.81    & \multicolumn{1}{c|}{51.06}    &  59.93    &  48.19   &  13.64   & 7.74  \\
                           & Reflexion              & 49.59    & 70.97    & 10.29    & \multicolumn{1}{c|}{42.65}    & 66.07     & 50.99    &    18.39  & 11.03    &  51.72   &   78.62  & 11.90    & \multicolumn{1}{c|}{42.86}    &  61.29    & 54.87    & 22.81    & 12.52                             
                  \\ \midrule
\multirow{3}{*}{GPT-4o} & CoT                    & 76.19    & 85.71    & 26.71    & \multicolumn{1}{c|}{70.59}    & 66.21     &  54.69    &  23.89    &  9.93   &  74.67   &  80.67   &  19.05   & \multicolumn{1}{c|}{63.81}    &  61.95    &  52.85    &  14.88   &  9.52   \\
                           & ReAct                  & 73.78    &  \underline{86.67}   &  17.65   & \multicolumn{1}{c|}{72.09}    &  66.91    &  56.43    &  17.27    &  13.97   &  75.27   &  \underline{82.95}   &  25.93   & \multicolumn{1}{c|}{62.96}    & 62.69     &  52.79    & 11.82    &  10.04   \\
                           & Reflexion              & \underline{76.52}    & 83.74  & \underline{27.94}  & \multicolumn{1}{c|}{\underline{75.12}}    &  \underline{67.05}    &  \underline{64.32}     &   \underline{34.56}     &  \underline{22.43}   & \underline{77.59}    & 79.31    &  \underline{26.28}   & \multicolumn{1}{c|}{\underline{75.07}}    &  \underline{63.01}    & \underline{63.73}     & \underline{27.97}    &   \underline{23.21}       
                           \\ \midrule
\multicolumn{2}{c|}{\textbf{DiMA}}   & \textbf{97.96}    & \textbf{96.94}   & \textbf{94.11}    & \multicolumn{1}{c|}{\textbf{91.18}}    &  \textbf{84.05}    & \textbf{97.89}     &  \textbf{58.83}    & \textbf{94.49}    & \textbf{98.82}   & \textbf{99.19}    &  \textbf{96.24}   & \multicolumn{1}{c|}{\textbf{97.62}}    &    \textbf{84.34}  & \textbf{97.36}     &  \textbf{63.69}   &  \textbf{93.45}  
\\ \midrule

\end{tabular}

}
\end{table*}

\subsection{Experimental Setup}
\subsubsection{Data Description}

We collect data from both the online environment and the role-playing dialog simulator constructed in section \ref{role-playing environment}. All datasets are collected from DiDi’s ride-hailing platform, covering the period from May 06, 2024, to August 24, 2024.
For both online and simulation-based evaluations, we utilize data from May 06, 2024, to May 24, 2024, as the training set for constructing the ride-hailing assistant. We select data from May 25, 2024, to July 25, 2024, as the testing set for the online evaluation. 
For offline evaluation, we use data from July 26, 2024, to August 09, 2024, to synthesize the testing set for Shanghai, and data from August 10, 2024, to August 24, 2024, to synthesize the testing set for Beijing.
The statistics of the three datasets are reported in Table~\ref{dataset}.

\subsubsection{Metrics}
We use four metrics for evaluation.
For the order planning task, we use round-level order accuracy (ROA) and session-level order accuracy (SOA) for evaluation. 
In particular, the ROA is defined as the ratio of dialog round with accurate order information, which satisfy user's request at current round.
The SOA measures the ratio of dialog sessions in which the order information is accurate for every dialog round throughout the entire session.
Similarly, we construct round-level response accuracy (RRA) and session-level response accuracy (SRA) for the response generation task.

\subsubsection{Baselines}
We compare our proposed framework with the following prevailing agent frameworks, including 
CoT \cite{wei2022chain} which introduces reasoning chain to enhance LLM's reasoning capacity, ReAct \cite{yao2023react} which synergizes reasoning and acting of LLM to solve given task, and Reflexion \cite{shinn2024reflexion} which is a multi-agent framework utilizing a self-reflection mechanism to iteratively enhance LLM's decision-making ability.
We select Qwen-2-72B and GPT-4o as the backbone of the above agent frameworks, aiming to demonstrate the effectiveness of DiMA against current state-of-the-art agent frameworks for ride-hailing.

\subsubsection{Evaluation Protocol}
In this work, we use both human evaluation and GPT self-evaluation.
\textbf{Human Evaluation.}
We employ an annotation team including 7 experts, each of them has a bachelor's degree or above in an IT-related field and possess work experience as an algorithm engineer at AI companies.
Before engaging in annotation work, they received more than one week of ride-hailing annotation training.
This training helps them how to learn to evaluate the correctness of trip orders and assistant responses according to serious product requirements.
We provide each annotator with an average hourly wage of 7.25 dollars.
Each record will be annotated by an annotator, and then another quality inspector will double-check whether the labeling is correct.
\textbf{GPT Evaluation.}
We also propose an LLM-based evaluation method, which is widely adopted in many recent studies \cite{zheng2023judging}.
In this work, we use GPT-4o as the evaluator.
For the trip order, we prompt GPT-4o to judge if the current order information is TRUE or FALSE.
For the assistant response, we prompt GPT-4o to provide a score (from 0 to 5), and the response with a score of more than 4 will be considered as TRUE.
The GPT evaluation method has also been used for annotation team to make a pre-labeling.

To reduce labor cost, we use GPT-4o to evaluate the model performance on the entire dataset and randomly sample 25\% data for human evaluation.
We provide more details in Appendix \ref{Evaluation of DiMA}.

\subsection{Online Test (RQ1)}
Table \ref{online results} presents the overall result of the proposed method across four evaluation metrics under six types of user conversation goals.
Under the human evaluation, it can be observed that DiMA achieves (93.83\%, 92.17\%, 87.11\%, 85.02\%) accuracy on round-level order creation, round-level response generation, session-level order creation, session-level response generation, respectively.
The GPT evaluation results (86.94\%, 94.66\%, 70.32\%, 86.68\%) also show the high accuracy of the constructed ride-hailing assistant, demonstrating its effectiveness.
In addition, we also find that order planning query accounts for the highest proportion (89.26\%), while DiMA achieves the best performance on casual chit-chat.
The performance variations across six types of user queries also indicates that distinct conversational goals may vary in complexity.

\subsection{Offline Test (RQ2)}
We also conduct experiments on two synthetic dataset to evaluate the effectiveness of DiMA compared with state-of-the-art agent frameworks.
The performance result is reported in Table \ref{simulation results}.
As can be seen from GPT evaluation results, DiMA achieves (25.35\%, 52.19\%, 70.23\%, 321.27\%) improvement compared with the state-of-the-art baseline using ROA, RRA, SOA, and SRA in Beijing.
The improvements in Shanghai are (31.88\%, 52.77\%, 128.71\%, 302.63\%), respectively.
In addition, DiMA has achieved significant improvement through human evaluation.
Specifically, the improvement on four metric are (28.02\%, 11.85\%, 236.83\%, 213.79\%) in Beijing, and (27.36\%, 19.58\%, 266.21\%, 30.04\%) in Shanghai, respectively.
More importantly, it is worth mentioning that the state-of-the-art multi-agent framework Reflexion performs poorly on order planning and response generation tasks, despite utilizing the powerful GPT-4o LLM backbone.
Furthermore, Reflexion significantly outperforms single-agent frameworks (\ie CoT and ReAct) in response generation tasks under GPT-evaluation, while the improvement in order planning task is relatively modest.
This may suggest that the complexity of these two tasks is different from each other, and the self-reflection mechanism may only be able to improve the response generation task.
Finally, we observe that all models achieves better performance on round-level metric (ROA and RRA) but worse on session-level metric (SOA and SRA), indicating the complexity of user queries may vary between different rounds.

\subsection{Ablation Study (RQ3)}
To validate the effectiveness of each module in DiMA, we conduct an ablation study on the Shanghai dataset using four metrics.
Specifically, we compare the following variants.
(1) DiMA w/o-TT removes the \textbf{T}ime \textbf{T}ool utilization in temporal reasoning process.
(2) DiMA w/o-CG removes designing different structured function for each \textbf{C}onversation \textbf{G}oal in progressive order planning module.
(3) DiMA w/o-MR removes the \textbf{M}ulti-type \textbf{R}eplier design.
(4) DiMA w/o-CLC removes the \textbf{C}ost-aware \textbf{L}LM \textbf{C}onfiguration, and the response generation task is randomly allocated.
As DiMA w/o-MR and DiMA w/o-CLC variants only affect response generation, we did not include their performance on the order planning.

\begin{table}[]
\caption{Ablation study on Shanghai dataset.}
\label{ablation study results}
\scalebox{0.8}{
\begin{tabular}{c|cccc|cccc}
\midrule
\multirow{2}{*}{Model Variants} & \multicolumn{4}{c|}{Human Evaluation (\%)} & \multicolumn{4}{c}{GPT Evaluation($\%$)} \\
                                & ROA   & RRA & SOA  & SRA  & ROA  & RRA & SOA & SRA \\ \midrule
DiMA w/o-TT                     &    88.29   &  91.03    & 90.48     & 95.85     & 76.71     & 88.37    & 60.69    & 90.23    \\
DiMA w/o-CG                     &  66.32     &  67.03    & 23.28     & 40.16     &   56.24   &   78.07   &  20.24   & 53.42    \\
DiMA w/o-MR                      & -      &  92.28    & -  &   95.24  &  -    &  92.72    &   -   & 80.35      \\
DiMA w/o-CLC                     &  -     &  87.26    & - &      79.05 &   -   &   76.54   & -     & 70.13    \\ 
\midrule
\textbf{DiMA}                            &  98.82     &  99.19    &   96.24   &  97.62    &     84.34 &   97.36   &  63.69   & 93.45    \\ \midrule
\end{tabular}
}
\vspace{-10pt}
\end{table}
As reported in Table \ref{ablation study results}, we obtain the following observations.
First, the time tool utilization contributes to the performance of the order planning task.
Removing it will decrease DiMA's accuracy on order planning tasks.
Second, after removing tailored function call for each conversation goal, the order accuracy decreases significantly, indicating the importance of progressive order planning module.
In addition, improper order planning will also lead to serious performance degradation on subsequent response generation tasks.
Third, we observe a performance improvement by defining three types of dialog repliers, which validates diverse response scenarios in ride-hailing assistant and the necessity of multi-type repliers.
Moreover, the cost-aware LLM configuration is very important, as we can observe significant performance degradation after removing explicit model and size allocation.

In addition, we also investigate the cross-city transferability. 
As shown in Table \ref{cross-city transferability}, we adopted two experimental setups: (1) training on Beijing data and testing on Shanghai data, and (2) training on Shanghai data and testing on Beijing data. 
The experimental results demonstrate that the LLM-based ride-hailing assistant is not influenced by city-specific data, exhibiting strong scalability.

\subsection{Continual Fine-Tuning Analysis (RQ4)}
Finally, we investigate the performance variation when employing continual fine-tuning.
We periodically monitoring system performance on real-world user data during the continual fine-tuning process, where new synthetic and real-user data are added to the fine-tuning dataset. 
Specifically, we evaluated the model on real-user interactions from May 9th to May 24th in our online system every 3 days. 
As shown in Figure \ref{fig:performance-vs-time}, continual fine-tuning with the augmented dataset (both synthetic and real-user samples, not synthetic alone) resulted in stable and consistent performance improvements throughout this period across all four metrics, without degrading its existing performance. This confirms that the simulator-generated data effectively complements real-world samples and supports the model's adaptation to diverse, realistic, and evolving user behavior patterns.
We provide data statistics over time in Appendix \ref{simulation}, and an case study in Appendix \ref{case study} for more illustrative analysis.

\subsection{Cost-Effectiveness Analysis (RQ5)}
\begin{figure}
    \centering
    \includegraphics[width=1 \linewidth]{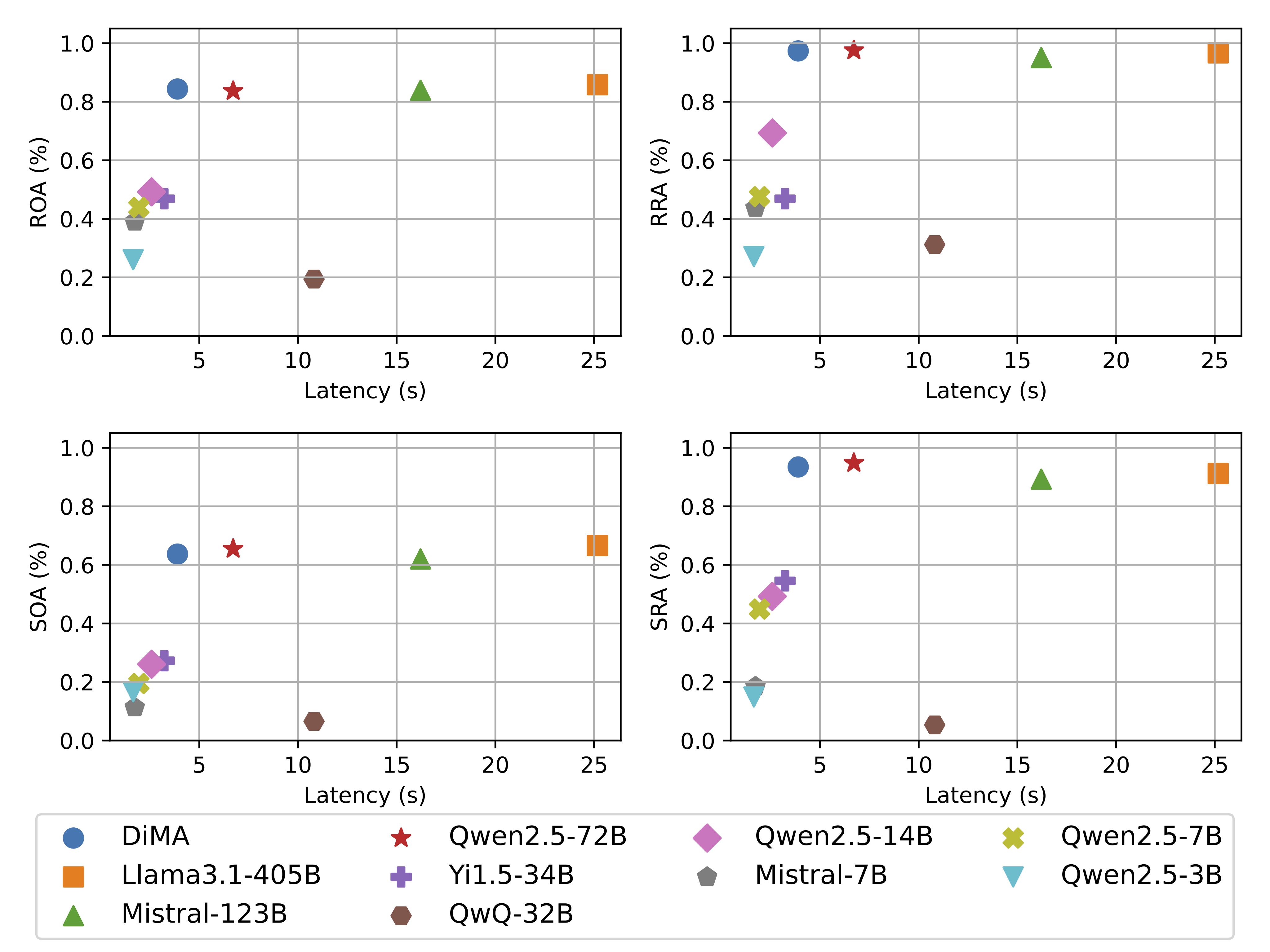}
    \caption{Comparison of performance and latency between DiMA and replacing the backbone of each of DiMA's modules as size-varying LLMs.
    Result is obtained on shanghai dataset.
    }
    \label{fig:performance-vs-latency}
\vspace{-10pt}
\end{figure}

We further discuss the cost-effectiveness of DiMA.
Specifically, we replace the backbone of every module in DiMA with size-varying LLMs to analyze potential variants in accuracy and latency.
As shown in Figure \ref{fig:performance-vs-latency}, we select, Llama3.1-405B, Mistral-123B \cite{Jiang2023Mistral7}, Qwen2.5-72B, Yi-1.5-34B \cite{young2024yi}, Qwen2.5-32/14/7B \cite{bai2023qwen}, Mistral-7B and QwQ-32B \cite{qwq-32b-preview} to replace the module backbones used in DiMA.
It can be observed that DiMA achieves comparable performance compared with replacing backbone with much larger LLM, while the inference latency was maintained at 3.89 seconds. 
While Qwen2.5-72B, Mistral-123B, and Llama3.1-405B could outperform DiMA by a small margin, they spend an additional amount of latency, respectively 0.72 times (3.89s->6.72s), 3.17 times (3.89s->16.21s) and 5.47 times more (3.89s->25.17s).
This renders them suboptimal for deployment in an online serving environment.
Additionally, DiMA's latency is higher than smaller models like Yi1.5-34B, Qwen2.5-14B, and Mistral-7B, but their performance across four metrics is significantly lower, reducing their competitiveness. 
Notably, the recently proposed large reasoning model (e.g., QwQ-32B \cite{qwq-32b-preview}) failed to show significant advantages, primarily due to its poor instruction-following ability and high reasoning latency.
Overall, the results demonstrate the advantages of DiMA's cost-aware LLM configuration mechanism in deployment.
\begin{table}[]
\caption{Study of cross-city transferability of DiMA on Beijing (BJ) and Shanghai (SH).}
\label{cross-city transferability}
\scalebox{0.8}{
\begin{tabular}{c|cccc|cccc}
\midrule
\multirow{2}{*}{Setting} & \multicolumn{4}{c|}{Human Evaluation (\%)} & \multicolumn{4}{c}{GPT Evaluation($\%$)} \\
                                & ROA   & RRA & SOA  & SRA  & ROA  & RRA & SOA & SRA \\ \midrule

\textbf{BJ $\to$ SH}                            &  98.03     &  98.92    &   95.96   &  96.81    &     83.95 &   96.13   &  63.20   & 93.13    \\
\textbf{SH $\to$ BJ}                            &  97.73     &  96.36    &   93.26   &  90.03    &     83.36 &   97.21   &  59.09   & 94.32   
\\ \midrule
\end{tabular}
}
\end{table}

\section{Related Work}
This work builds upon task-oriented dialog systems and LLM-based assistant.
We review representative work in both areas below.

\subsection{Task-oriented Dialog Systems}
Task-oriented dialog (ToD) systems facilitate users in achieving specific objectives through natural language interactions, such as making reservations or inquiries \cite{qin2023end}. 
In recent years, numerous methodologies \cite{hosseini2020simple, he2022galaxy, yang2021ubar, hudevcek2023llms, chung2023instructtods} have been proposed to enhance ToD system performance. However, existing ToD frameworks lack specialization for open-world ride-hailing scenarios. The dynamic nature of ride-hailing interactions—spanning order creation, modification, and cancellation—necessitates a tailored solution that current ToD approaches fail to provide.

\begin{figure}
    \centering
    \includegraphics[width=1 \linewidth]{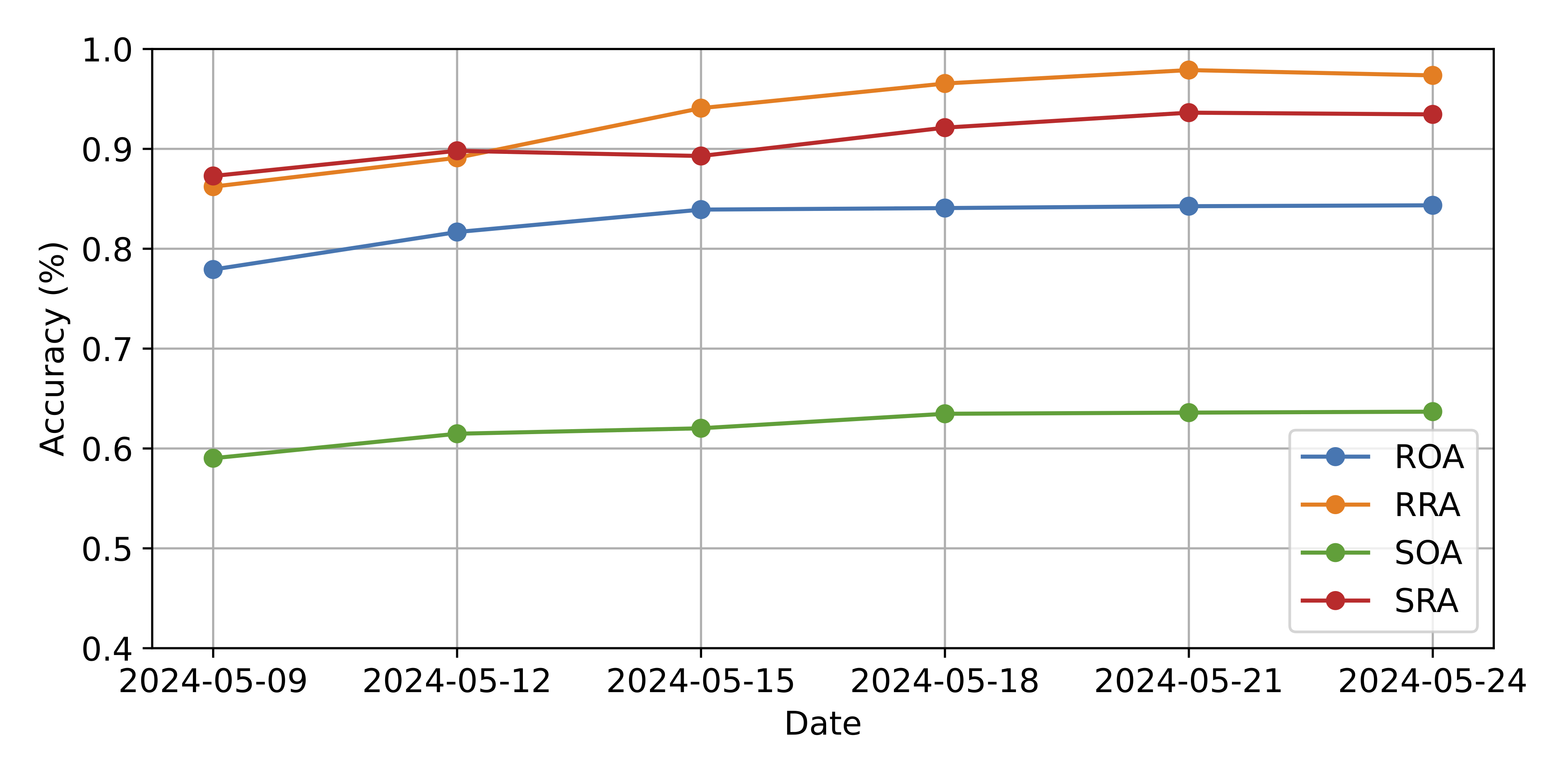}
    \caption{Comparison of DiMA's performance over the time. We periodically (every three days) fine-tune DiMA using collected data up to the latest date.}
    \label{fig:performance-vs-time}
    \vspace{-10pt}
\end{figure}

\subsection{LLM-based Assistants}
The development of domain-specific intelligent assistants powered by Large Language Models (LLMs) \cite{dong2023towards} has gained significant traction in both academia and industry. LLMs enable automated planning and execution of complex industrial tasks across various domains. For instance, Mind2Web \cite{deng2024mind2web}, WebGPT \cite{gur2023real}, and AutoWebGLM \cite{lai2024autowebglm} have been introduced as LLM-powered web navigation assistants. LLMPA \cite{guan2023intelligent} serves as a virtual assistant for travel planning within the Alipay App, while Flowris \cite{sun2023flowris} and GRILLBot \cite{fischer2024grillbot} function as intelligent assistants for data management and multimodal conversation, respectively. Despite these advancements, no LLM-based assistant has been designed specifically to automate ride-hailing services, highlighting a gap in existing research.

\section{Conclusion}
In this paper, we introduced DiMA, an LLM-powered ride-hailing assistant designed to enhance user experience and operational efficiency in real-world ride-hailing scenarios. DiMA integrates multiple key components, including a spatiotemporal-aware tool-augmented order planning module, which ensures accurate ride-hailing order creation by leveraging external spatial and temporal tools. Additionally, we developed a cost-effective dialog system, which dynamically allocates response tasks to cost-aware LLM repliers, optimizing the trade-off between response quality and system latency. A continual fine-tuning scheme has been applied to align the assistant's behavior with the human decision-making process.
The online and offline evaluation demonstrated the effectiveness of DiMA's performance in handling diverse ride-hailing user requests, achieving high accuracy and efficiency. 
The assistant has been deployed on DiDi's mobile application since May 2024.

\begin{acks}
This work was supported by the National Key R\&D Program of China (Grant No.2023YFF0725004), National Natural Science Foundation of China (Grant No.92370204), the Guangzhou Basic and Applied Basic Research Program under Grant No. 2024A04J3279, Education Bureau of Guangzhou Municipality. 
We appreciate the guidance provided by the engineers in Didichuxing Co. Ltd.
\end{acks}

\bibliographystyle{ACM-Reference-Format}
\balance
\bibliography{sample-base}

\clearpage

\appendix
\section{Appendix}
\label{Appendix:NULL_APPENDIX}
\subsection{Simulation Environment Construction}
\label{simulation}
We provide more details here on how to use collected online data for simulation testing, including dialog content generation, GPT-based user role-playing, and the instruction set statistics information.

\subsubsection{Dialog Content Generation}
Given a collected online data, we prompt GPT-4o to generate a dialog content based on a randomly generated start location and end location, and randomly generated user profile.
Specifically, the online data is a multi-dialog content between user and ride-hailing assistant, \ie <user: Hi, book a order to Tsinghua for me, assistant: Sure, user: ..., assistant: ...>.
To save the personal privacy, any semantic location information (\eg phone number) related to the user are removed.
The randomly generated start/end location contains a address name, latitude and longitude, \eg \emph{<Dongli Garden-Side Gate, 31.1862, 121.5165>}.
The randomly generated user profile contains user's occupation and age, \eg \emph{<Lawyer, 38 years old>}.

With the above information, we prompt GPT-4o to follow the real-world multi-dialog content style to generate a simulated dialog content, and then output a ride-hailing intent and the dialog timestamp.
We provide the generated example in the Figure \ref{Dialog Content}.

\begin{figure}
    \centering
\begin{lstlisting}
Dialog Timestamp: 2024-7-28 19:00:00

Ride-hailing Intent: The user wants to travel to the intersection of Lou Shan Guan Road and Mao Tai Road.

Simulated Dialog Content:
<User>: I want to go to the intersection of Lou Shan Guan Road and Mao Tai Road.
<Assistant>: Okay, XiaoDi has set "the intersection of Lou Shan Guan Road and Mao Tai Road" as your drop-off point. The driver is expected to pick you up between 19:58 and 20:02 today, and it will take approximately 25 minutes to reach your destination.
<User>: Will there be traffic during this time period?
<Assistant>: Sorry, XiaoDi is unable to provide real-time traffic flow information.
\end{lstlisting}
    \caption{An illustrative example of generated dialog content in Shanghai dataset.}
     \label{Dialog Content}
\end{figure}

\subsubsection{GPT-based User Role-playing}
With the generated dialog content using collected online data, we can further prompt GPT-4o to play the role of a real-world user following the dialog content. Specifically, we first provide a task profile in the prompt to guide GPT-4o in understanding the task, and then we provide the dialog content and prompt it to generate the first sentence to begin a dialog. 
In such a multi-dialog scenario, the response of the assistant will also be iteratively fed into the prompt to help GPT-4o understand if the ride-hailing intent has been completed.
We provide detailed prompt in our GitHub project \href{https://github.com/usail-hkust/DiMA}{https://github.com/usail-hkust/DiMA} to facilitate the understanding of GPT-based user role-playing.

\subsubsection{Instruction Dataset Statistics}
As can be seen in Table \ref{instruction set statistics}, we report the statistic information of mixed instruction set used for continual fine-tuning.
Overall, origin the training set (mixed instruction set from May 06 2024 to May 24 2024) contains 2,086 user queries.
After data augmentation with role-playing simulator, the instruction set is expanded, \eg the instruction set quantities of order planning and dialog replier selection is 3,649 and 3,927 respectively. 
For three types of replier, we observe that DiMA allocates specialized replier for response generation in most cases.
This resulted in a smaller instruction set for the other two repliers, \ie the instruction set quantities of error-handling replier and knowledge-enhanced replier is 458 and 475 respectively. 
The performance degradation caused by fine-tuning may also stem from this reason. 
Therefore, as discussed in Section \ref{deployment}, we did not fine-tune LLMs for error-handling replier and knowledge-enhanced replier  during actual deployment.
Moreover, we also display the instruction tuning dataset statistics (used in Figure \ref{fig:performance-vs-time}) over the time to facilitate better understanding of continual fine-tuning mechanism.

\begin{table*}[]
\centering
\setlength{\abovecaptionskip}{-0.05cm}
\caption{The augmented instruction set for every module in DiMA.}
\label{instruction set statistics}
\scalebox{0.85}{
\begin{tabular}{c|cc|ccccc}
\midrule
\multirow{3}{*}{Time span} & \multirow{3}{*}{Users} & \multirow{3}{*}{User queries} & \multicolumn{5}{c}{Mixed instruction set}                                                                                                                                                                                                                                                         \\
                           &                        &                               &\begin{tabular}[c]{@{}c@{}}Order \\ planning\end{tabular} & \begin{tabular}[c]{@{}c@{}}Dialog replier \\ selection\end{tabular} & \begin{tabular}[c]{@{}c@{}}Specialized \\ replier\end{tabular} & \begin{tabular}[c]{@{}c@{}}Error-handling \\ replier\end{tabular} & \begin{tabular}[c]{@{}c@{}}Knowledge-enhanced \\ replier\end{tabular} \\ \midrule
2024/05/06 - 2024/05/09    & 160                    & 508                           & 837            & 849                                                                 & 572                                                            & 102                                                               & 115                                                                   \\
2024/05/09 - 2024/05/12    & 142                    & 438                           & 762            & 811                                                                 & 513                                                            & 76                                                                & 109                                                                   \\
2024/05/12 - 2024/05/15    & 162                    & 461                           & 771            & 906                                                                 & 522                                                            & 94                                                                & 93                                                                    \\
2024/05/15 - 2024/05/18    & 50                     & 158                           & 303            & 312                                                                 & 208                                                            & 42                                                                & 39                                                                    \\
2024/05/18 - 2024/05/21    & 67                     & 210                           & 412            & 422                                                                 & 273                                                            & 62                                                                & 45                                                                    \\
2024/05/21 - 2024/05/24    & 101                    & 311                           & 564            & 627                                                                 & 368                                                            & 82                                                                & 74                                                                    \\ \midrule
Overall                    & 682                    & 2,086                          & 3,649           & 3,927                                                                & 2,456                                                           & 458                                                               & 475                                                                   \\ \midrule
\end{tabular}
}
\end{table*}

\subsection{Prompt Template of DiMA}
\label{prompt_template}
This section presents the prompt template of every module in DiMA to facilitate a better understanding.
We release the prompt template of the following five components in our GitHub project \href{https://github.com/usail-hkust/DiMA}{https://github.com/usail-hkust/DiMA}: (1) Spatiotemporal-aware Tool-augmented Order Planning; (2) Multi-type Dialog Replier Selection; (3) Specialized Dialog Relpier; (4) Error-handling Dialog Replier; (5) Knowledge-enhanced Dialog Replier.

\subsection{Evaluation of DiMA}
\label{Evaluation of DiMA}

\begin{table*}[]
\caption{An illustrative evaluation forum.}
\label{evaluation form}
\scalebox{0.85}{
\begin{tabular}{cccccc|ccc}
\midrule
\multicolumn{6}{c|}{Sample}                                & \multicolumn{2}{c}{Evaluation} & \multirow{2}{*}{Validation}\\
Time & Session\_id & Round\_id & User query & Order & Response & Order        & Response  &        \\ \midrule
   2024-07-01 13:23:25 &   XXXX         &  0         &  To Peiking University now    & <Order Info>      &  \begin{tabular}[c]{@{}c@{}}Sure. Click the 'Confirm'\\ button to place this order.\end{tabular}      &   TRUE           & TRUE & TRUE                \\
   ...  &  ...           &   ...        &  ...    &  ...     & ...         &   ...           &    ... &    ...            \\ \midrule
\end{tabular}}
\end{table*}
We evaluate the performance of ride-hailing assistant from two perspective, \ie order planning and assistant response.
All prompt template could be found in our GitHub project. 

For order planning, we explicitly prompt the LLM to return its evaluation results and corresponding reasons. Specifically, we first clearly explain each of conversation goals, \eg confirming the order or wanting to cancel the order. Then we explain what information should be contained in the order information, \eg start location and the price of the car type. After that, we prompt GPT-4o to evaluate if the current order information satisfies the user's intent and to return its evaluation.
Overall, in the GPT evaluation, the following spatiotemporal information and order information should be evaluated: (1) start location; (2) via location; (3) end location; (4) departure time; (5) arrival time; (6) price and car type in the order information; (7) user intent. Only if all the above information is correct will the LLM view the order creation as TRUE.

For the response generation, we first provide several evaluation examples in the prompt (both positive and negative samples) to cover three response scenarios.
Then we provide evaluation rules, which is mainly formulated by the product manager for DiDi ride-hailing assistant, to guide LLM evaluate if current response can satisfy user's request and align with pre-defined product rule.
Specifically, the LLM will return a score (from 0 to 5), the detailed scoring criteria is as follow:
\begin{itemize}
    \item Scored 1 out of 5: If the response is completely irrelevant to the user's conversation, give 1 point.
    \item Scored 2 out of 5: If the response contains incorrect information, is inconsistent with the basic information provided by <executed results> and <order information>, promises features not supported by <executed results> or <order information>, or fails to address abnormal content in either of the two, give 2 points.
    \item Scored 3 out of 5: If the response is relevant and provides some information related to the user's inquiry, but there are issues with phrasing and a significant amount of redundant content, give 3 points.
    \item Scored 4 out of 5: If the response directly and comprehensively addresses the user's question but has some room for improvement in phrasing, give 4 points.
    \item Scored 5 out of 5: If the response perfectly addresses the user's question with appropriate phrasing, is concise and polite, give 5 points.
\end{itemize}

\subsubsection{Detailed Process of Human Evaluation}
The human evaluation process consists of two sequential steps, \ie annotation and quality validation.
As shown in Table \ref{evaluation form}, in the first step, we ask human annotator to fill the evaluation forum.
Specifically, given the user query, created order and generated response, the annotator will follow serious annotation rules created by product manager at DiDi to evaluate the results.
In the second step, another inspector will be required to double-check if these evaluation is true.
Once such a evaluation process is finished, we will calculate the overall evaluation metric, \ie the round-level order accuracy (ROA) and session-level order accuracy (SOA), round-level response accuracy (RRA) and session-level response accuracy (SRA).

\subsection{Case Study of DiMA}
\label{case study}
We present a case study of DiMA to illustrate how DiMA creates a ride-hailing order and provides a response given a user query. 
The figure shown in our GitHub Project \href{https://github.com/usail-hkust/DiMA}{https://github.com/usail-hkust/DiMA} shows the entire dialog session. 

As can be seen, in the first dialog round, the user said "Hello, Xiaodi. I want to go to Huateng Garden South-Gate". 
Then we concatenate this user query and the current time into a prompt, which is fed into the spatiotemporal-aware tool-augmented order planning module. 
Based on the generated function, DiMA will automatically invoke APIs and tools to execute these functions and obtain the executed results and corresponding information. 
In this ride-hailing scenario, there are multiple candidate drop-off locations. 
Therefore, DiMA allocate this task to a specialized dialog replier. 
As can be seen, the replier successfully guides the user to select his preferred drop-off location.

In dialog round 2, we first concatenate history user query, function call, response into the prompt and then DiMA generates function call for this round.
After executing generated function, we obtain the ride-hailing order information and the executed results.
However, current user request (want to book a order less than 40 dollars) cannot be satisfied because all available order are priced greater than 40 dollars.
Therefore, DiMA allocates this response task for error-handling dialog replier, who first guides the user to confirm this suggested order, then politely rejects the user's infeasible prices requirement and provides corresponding explanation. 
After such guidance, the user follows the suggestion to click the "Confirm" button.

In the last dialog round, the user wants to ask about information on nearby restaurants and even hopes the assistant can book a restaurant. 
This scenario involves ride-hailing policies, thus DiMA allocates this response task to knowledge-enhanced dialog replier. 
According to the retrieved policy information, this is not within the scope of the assistant's duties, thus the replier tactfully rejects the user's request and make an apology.
\end{document}